\documentclass{llncs}
\usepackage{cleveref}
\usepackage{bbm}
\usepackage{amsmath}
\usepackage{amsfonts}
\usepackage{enumerate}
\usepackage{tikz}
\usetikzlibrary{fit,positioning}
\usepackage{scalefnt}
\usepackage{array}
\usepackage{subcaption}
\usepackage{etoolbox}
\abovedisplayskip
\belowdisplayskip
\tikzstyle{dcircle}=[circle, draw=black, minimum width = 1cm]
\tikzstyle{connect}=[-latex, thick]

\preto\equation{\par\nobreak\small\noindent}
\preto\align{\par\nobreak\small\noindent}
\preto\eqnarray{\par\nobreak\small\noindent}
\expandafter\preto\csname align*\endcsname{\par\nobreak\small\noindent}
\begin{document}


\title{Topic Model Based Multi-Label Classification from the Crowd}

\author{ Divya Padmanabhan \and Satyanath Bhat \and Shirish Shevade \and Y. Narahari}
\institute{Indian Institute of Science, Bangalore}
\date{}
\maketitle
\begin{abstract}
Multi-label classification is a common supervised machine learning  problem where each instance is associated with multiple classes. The key challenge in this problem is learning the correlations between the classes. 
An  additional challenge arises  when the labels of the training instances are provided by noisy, heterogeneous crowdworkers with unknown
qualities.  We first assume labels from a perfect source and propose
a novel topic model where the present as well as the absent classes generate  the  latent topics and hence the words. We non-trivially extend our topic  model to the  scenario where the labels are provided by noisy crowdworkers.   Extensive experimentation on real world datasets reveals the superior  performance  of the proposed model.  The proposed model learns the  qualities of the annotators as well, even with minimal training data.
\end{abstract}
\section{Introduction}
Multi label classification is a variant of a classification problem wherein an instance $\bold{d}$ is associated with multiple classes or labels. There are several areas where multi-label classification finds applications, for example, text classification, image retrieval, etc. Consider the task of classification of documents into several classes such as crime, politics, arts, sports etc. The classes are not mutually exhaustive since a document belonging to the `politics' category may also belong to `crime'. In the classification of images, an image belonging to `forest' category may also belong to `scenery' category, and so on. 

One of the solution approaches for multi-label classification is to generate a new label set that is a power set of the original label set, and then use traditional single label classification techniques.  The immediate limitation here is an exponential blow-up of the label set and availability of only a small sized training data for each of the generated labels. Another approach is to build one-vs-all binary classifiers, where, for each label, a binary classifier is built. This method, however, does not take into account the correlation between the labels.

In the past, topic models \cite{lda} have proved to be successful in modeling the process behind generating text documents. The idea is to model latent topics responsible for generating words. Originally, topic models were used in an unsupervised manner and were gradually adapted to the supervised learning setting. In the supervised setting, the topics and hence words are assumed to be generated depending only on the classes that are present. The topic models discussed in the literature do not make use of the information provided by `absence' of classes.  The absence of a class often provides critical information about the words present. For example, a document labeled `sports' is less likely to have words related to `arts'. Similarly in the images domain, an image labeled `city' is less likely to have the characteristics of `forest'. Needless to say, such correlations are dataset dependent. However a principled analysis must account for such correlations. Motivated by this subtle observation, we introduce a novel topic model for multi-label classification.

Further the problem renders itself more interesting when the labels are procured from multiple heterogenous noisy crowd-workers whose qualities are unknown. We also refer to crowd-workers as annotators in the paper. In the current era of big data where large amounts of unlabeled data are readily available, obtaining a noiseless source for labels is almost impossible. However it is possible to get instances labeled by several human annotators. The problem becomes harder as now the true labels are unknown and the qualities of the annotators must be learnt in order to train a model. We non-trivially extend our topic model to this scenario.

\subsubsection*{Contributions}
\begin{enumerate}

\item We introduce a novel topic model  for multi-label classification; our model has the distinctive feature of exploiting any additional information provided by the absence of classes. We refer to our topic model as ML-PA-LDA (Multi-label Presence-Absence LDA).
\item If the labels are provided by multiple noisy annotators (from a crowd), we enhance our model to account for heterogenous annotators with  unknown qualities.
We refer to this enhanced model as ML-PA-LDA-C (ML-PA-LDA with Crowd). A feature of ML-PA-LDA-C is it does not require an annotator to label all classes for a document. Even partial labeling by the annotators upto the granularity of labels within a document is adequate.
\item Our experiments on real world datasets establish the superior performance of both of our models. The qualities of the annotators learned by our algorithm approximate closely the true qualities of the annotators.

\end{enumerate}
We now describe relevant approaches available in the literature.
\section{Related Work}
Several approaches have been devised for multilabel classification with labels provided by a single  source. The most natural approach is the Label Powerset (LP) method \cite{ChermanMM11} which generates a new class for every combination of labels and then solves the problem using multiclass classification approaches. The main drawback of this approach is the exponential growth in the number of classes, leading to several generated classes having very few labeled instances leading to overfitting. To overcome this drawback, RAndom k-labELsets method (RAkEL) \cite{RAkel} was introduced, which constructs an ensemble of LP classifiers where each classifier is trained with a random subset of $k$ labels. However, the large number of labels still poses challenges. The approach of pairwise comparisons (PW)  improves upon the above methods, by constructing C(C-1)/2 classifiers for every pair of classes, where $C$ is the number of classes. Finally a ranking of the predictions from each classifier yields the labels for a test instance. Rank-SVM \cite{ranksvm} uses PW approach to construct SVM classifiers for every pair of classes and then performs a ranking.

The previously described approaches are discriminative approaches. Generative models for multilabel classification model the correlation between the classes using mixing weights for the classes \cite{McCallum99multi-labeltext}. Other probabilistic mixture models include Parameteric Mixture Models PMM1 and PMM2 \cite{Ueda03parametricmixture}. After the advent of the topic models like Latent Dirichlet Allocation (LDA) \cite{lda}, extensions have been proposed for multi-label classification such as Wang et al
\cite{Wang-correlatingTopicModel}. However in  \cite{Wang-correlatingTopicModel}, due to the non-conjugacy of the distributions involved, closed form updates cannot be obtained for several parameters and iterative optimization algorithms such as conjugate gradient and Newton Raphson are required to be used in the variational E step as well as M step, introducing additional implementation issues. Adapting this model to the case of crowds would result in  enormous complexity. The topic models proposed for multi-label classification in \cite{Rubin2012} involve far too many parameters which can be learnt effectively only in the presence of large amounts of labeled data. For small and medium sized datasets, the approach suffers from overfitting. Moreover it is not clear how this model can be adapted when labels are procured from crowdworkers with unknown qualities. SLDA \cite{slda} is a single label classification technique which works well on multi-label classification when used with the one-vs-all approach. SLDA inherently captures the correlation between classes through the latent topics.

With crowdsourcing gaining popularity due to the availability of large amounts of unlabeled data and difficulty in procuring noiseless labels for these datasets, aggregating labels from the crowd has become an important problem. Raykar et al \cite{Raykar2010} look at training binary classification models with labels from a crowd with unknown annotator qualities. Being a model for multiclass classification, this model does not capture the correlation between classes and thereby cannot be used for multi-label classification from the crowd. 
Mausam et al \cite{BraggMW13} look at multi-label classification for taxonomy creation from the crowd. They construct $C$ classifiers by modeling the dependence between the classes explicitly. The graphical model representation involves too many edges especially when the number of classes is large and hence the model suffers from overfitting.
Deng et al \cite{Deng2014} look at selecting the instance to be given to a set of crowd-workers. However they do not look at aggregating these labels and developing a model for classification given these labels. In the report \cite{Duanleveraging}, Duan et al. look at methods to aggregate a multi-label set provided by crowd-workers. However, they do not look at building a model for classification for new test instances for which the labels are not provided by the crowd.  
Recently the topic model, SLDA, has been adapted to learning from the labels  provided by crowd annotators \cite{rodrigues2015learning}. However, like its predecessor SLDA, it is only applicable to the single label setting and not to multi-label classification.

The existing topic models in the literature assume that the presence of a class generates words pertaining to those classes and do not take into account the fact that the absence of a class may also play a role in  generating words. In practice, the absence of a class may yield information about occurrence of words. We propose a model for multi-label classification based on latent topics
where the presence as well as absence of a class generates topics. The labels could be procured from a set of crowd workers whose qualities are unknown. 
\section{Our Approach for Multi-label Classification: ML-PA-LDA}
\label{sec:multi-label-model}
We now explain our model for multi-label classification.  For ease of exposition, we use  notations from the text domain. However the model itself is general and can be applied to several domains by suitable transformation of features into words. In our experiments we have applied the model to domains other than text. We will explain the transformation of features to words when we describe our experiments.

Let $D$ be the number of documents in the training set, $C$ the total number of classes,  $T$ the number of topics and $K$ the number of annotators. In multi-label classification, a document may belong to any `subset' of the $C$ classes as opposed to the standard classification setting where a document belongs to exactly one class. We denote by $V$ the size of the vocabulary $\nu = \{\nu_1, \ldots, \nu_V \}$, where $\nu_j$ refers to the $j^{th}$ word in $\nu$. Consider a document $\bold{d}$ comprising $N$ words $\bold{w} = \{ w_1, w_2, \ldots, w_N\}$ from the vocabulary $\nu$.  Let $\lambda = \left[\lambda_1, \ldots, \lambda_C\right] \in \{0,1\}^C$ denote the true class membership of the document. In our notations, we denote by $w_{nj}$ the value $\mathbbm{1}[w_n = \nu_j]$, that is the indicator that the word $w_n$ is the $j^{th}$ word of the vocabulary. Similarly, we denote by $\lambda_{ij}$, the indicator that $\lambda_i = j $, where $j=0 $ or $ 1$. The objective is to obtain $\lambda$ for every test document. 

\subsection*{Topic Model for the Documents}
We introduce a model to capture the correlation between the various classes generating a given document. The presence as well as absence of a class provides additional information about the topics present in a document. 
We now describe the generative process for each document assuming labels are provided by a perfect source.
\begin{enumerate}
\item Draw  $\lambda_i \sim \text{Bern}(\xi_i)$ for every class $i=1, \ldots, C$.
\item Draw $\theta_{i,j,.} \sim \text{Dir}(\alpha_{i,j,.})$ for $i=1, \ldots, C$, for $j \in \{0,1\}$, where $\alpha_{i,j,.}$ is a Dirichlet distribution with $T$ parameters.
\item For every word $w$ in the document
\begin{enumerate}
\item Sample $u \sim \text{Unif}\{1, \ldots, C\}$ from one of the $C$ classes.
\item Generate a topic $z \sim \text{Mult}(\theta_{u,\lambda_u,.})$,  where $\theta_{i,j,.}$ is a multinomial distribution in $T$ dimensions.
\item Generate the word $w \sim \text{Mult}(\beta_{z.})$ where $\beta_{t.}$ is a multinomial distribution in $V$ dimensions.
\end{enumerate}
\end{enumerate}
We refer to this model where the true class vector $\lambda$ is observed for the training documents as ML-PA-LDA (Multi-label Presence-Absence LDA).

\subsection*{Single Coin Model for the Annotators} When the true labels of the documents are not observed, $\lambda$ is unknown. Instead noisy versions $y_1, \ldots, y_K$ of $\lambda$ provided by a set of $K$ independent annotators with heterogenous unknown qualities $\{ \rho_1, \ldots, \rho_K \}$ are observed.$y_{ji}$ can be either $ 0, 1 $ or $-1$. $y_{ji} = 1$ indicates that, according to annotator $j$, the class $i$ is present while  $y_{ji} = 0$ indicates that the class $i$ is absent as per annotator $j$. $y_{ji}= -1$ indicates that the annotator $j$ has not made a judgement on the presence of class $i$ in the document. This allows for partial labeling upto the granularity of labels even within a document. This flexibility in the modeling is essential, especially when the number of classes is large.
$\rho_j$ is the probability with which an annotator reports the ground truth corresponding to each of the classes. $\rho_j$ is not known to the learning algorithm. For simplicity we have assumed the single coin model for annotators and also that the qualities of the annotators are independent of the class under consideration. That is, $P(y_{j1} = 1| \lambda_1=1) = P(y_{j1} = 0| \lambda_1=0) = \ldots = P(y_{jC} = 1| \lambda_C=1) = P(y_{jC} = 0| \lambda_C=0) = \rho_j$.

\begin{figure}
\begin{minipage}{0.6\linewidth}
\includegraphics[scale=0.55]{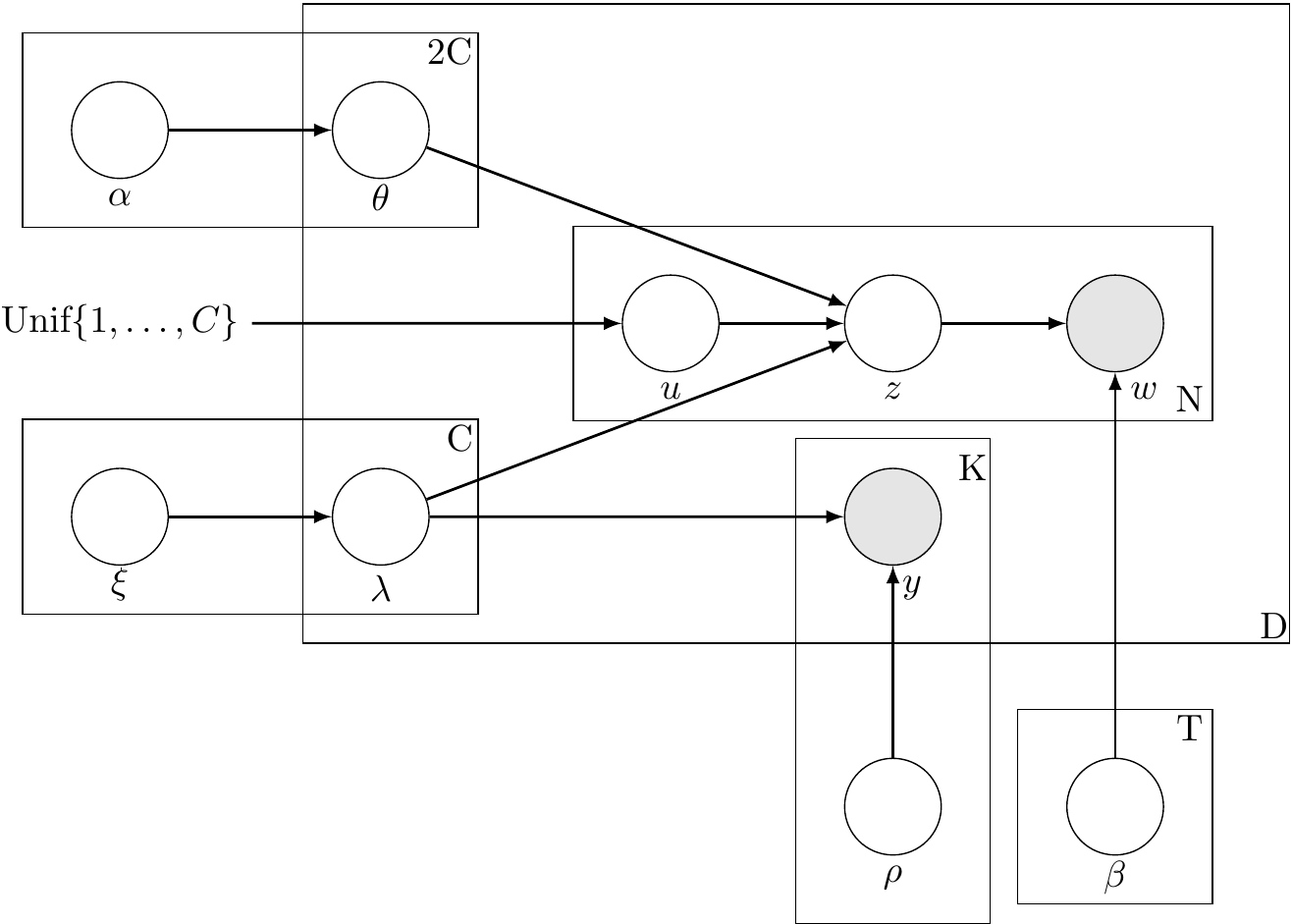}
\subcaption{Graphical model for ML-PA-LDA-C where labels are provided by the crowdworkers. }
\label{fig:graph-model}
\end{minipage}
\begin{minipage}{0.39\linewidth}
\includegraphics[scale=0.44]{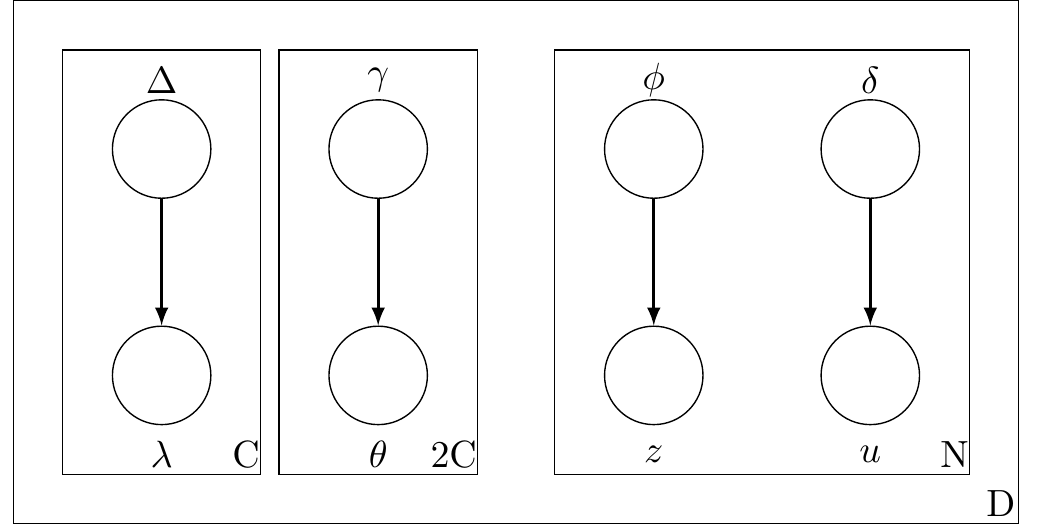}
\subcaption{Graphical model representation of the variational distribution used to approximate the posterior }
\label{fig:var_model_non_smooth}
\end{minipage}
\caption{Graphical model representations of our model. For ML-PA-LDA (the non-crowd version), \Cref{fig:graph-model} is modified so that the observed variables are $w$ and $\lambda$ while the random variable $y$ and therefore the parameter $\rho$ is absent.}
\label{fig:main_model}
\end{figure}

The generative process for the documents is depicted pictorially in Figure \ref{fig:graph-model}. The parameters of our model consist of 
  $\pi= \{\alpha$ ,  $\xi$ ,  $\rho$,   $\beta \}$. 
The observed variables for each document are 
$\bold{d} = \{ \bold{w}, y_{ij} \}$ for $i = 1, \ldots, C$, $j = 1, \ldots, K$. The hidden random variables are $ \Theta= \{ \theta, \lambda, u, z \}$. We refer to our topic model trained with labels from annotators as ML-PA-LDA-C (multi-label presence-absence LDA from crowds).
\section{Variational EM for ML-PA-LDA-C}
We now detail the steps for estimating the parameters of our proposed model. Since ML-PA-LDA-C is a generalization of ML-PA-LDA to adapt to the crowd, we provide the details of the steps for ML-PA-LDA-C and give pointers to highlight the differences with ML-PA-LDA whenever appropriate.

Given the observed words $\bold{w}$ and the labels $y_1, \ldots, y_k$ for a document $\bold{d}$, the objective of the model described above is to obtain $p(\Theta|\bold{d})$. Here, the challenge lies in the intractable computation of  $p(\Theta|\bold{d})$  which arises due to the intractability in the computation of $ p(\bold{d}|\pi)$. We use variational inference with mean field assumptions to overcome this challenge.

Suppose $q( \Theta)$ is any distribution over $\Theta$
for any arbitrary $\Theta = \{ \theta, \lambda, u, z \}$ which approximates $p(\Theta|\bold{d})$.
The underlying variational model is provided in \Cref{fig:var_model_non_smooth}.

\begin{align}
\label{eqn:elbo}
\ln p(\bold{d}|\pi) &= \ln \frac{p(\bold{d},\Theta)}{p(\Theta|\bold{d})} = \ln \frac{p(\bold{d},\Theta) q( \Theta)}{q( \Theta)p(\Theta|\bold{d}) } = \mathbb{E}_{q(\Theta)} \left[\ln \frac{p(\bold{d},\Theta) q( \Theta)}{q( \Theta)p(\Theta|\bold{d}) } \right] \nonumber \\
&=\mathbb{E}_{q(\Theta)}\left[\ln p(\bold{d},\Theta) - \ln q( \Theta)\right] + \mathbb{E}_{q(\Theta)}\left[\ln q(\Theta)- p(\bold{d}|\Theta)\right] \nonumber \\
&= \mathcal{L}(\Theta) + \textbf{KL}(q(\Theta) ||p(\Theta|\bold{d})
\end{align}
The idea is to maximise $\mathcal{L}(\Theta)$ over the variational parameters $\{ \delta, \gamma, \phi, \delta \}$ so that $\textbf{KL}(q(\Theta) ||p(\bold{d}|\Theta)$ also gets minimized.
\begin{align*}
\ln & p(\bold{d},\Theta | \pi)= \ln p(\bold{w}, y, \theta, \lambda, u, z | \pi) \nonumber\\
& = \ln p(\lambda| \xi) + \ln p(\theta | \alpha) + \ln p(u) + \ln(z|\lambda, u, \theta) + \ln p(\bold{w}|z, \beta) + \ln p(y|\lambda, \rho)
\end{align*}
\begin{equation}
\label{eqn:lambda-pdf}
\ln p(\lambda|\xi) = \sum_{i=1}^C \lambda_i \ln \xi_i + (1 - \lambda_i) \ln (1- \xi_i)
\end{equation}
\begin{equation}
\label{eqn:theta-pdf}
\ln p(\theta_{ij}|\alpha) = \ln \Gamma \left(\sum_{t=1}^T \alpha_{ijt}\right) - \sum_{t=1}^T \ln \Gamma \alpha_{ijt}
+ \sum_{t=1}^T (\alpha_{ijt} - 1)\log \theta_{ijt}
\end{equation}
\begin{equation}
\label{eqn:u-pdf}
\log p(u) = \sum_{n=1}^N \sum_{i=1}^C u_{ni} \log 1/C
\end{equation}
\begin{equation}
\label{eqn:z-pdf}
\log p(z|u, \lambda, \theta) = \sum_{n=1}^{N} \sum_{t=1}^T\sum_{i=1}^C\sum_{j=0}^{1} u_{ni} \lambda_{ij} z_{nt} \log \theta_{ijt} 
\end{equation}
\begin{equation}
\label{eqn:w-pdf}
\log p(w|z, \beta) = \sum_{n=1}^N\sum_{t=1}^T\sum_{j=1}^V w_{nj} z_{nt} \log \beta_{tj}
\end{equation}
\begin{align}
\label{eqn:y-pdf}
\log p(y|\lambda, \rho) = &\sum_{j=1}^{K}\sum_{i=1}^C \left[\lambda_i y_{ij} + (1-\lambda_i)(1 - y_{ij})\right] \log \rho_j \nonumber \\&\;\;\;\; +
\left[ (1- \lambda_i) y_{ij} + \lambda_i(1 - y_{ij})\right] \log  1 - \rho_j
\end{align}

Assume the following variational distributions over $\Theta$ for a document $d$.\\
$u^d \sim \text{Mult}(\delta^d) \;\;\;\;\;\;\;$, 
$\lambda_i^d \sim \text{Bern}(\Delta_i^d)$ for $i= 1, \ldots, C\;\;\;\;\;\;\;$,$z^d \sim \text{ Mult}(\phi^d) $\\
$\theta_{ij}^d \sim \text{Dir}(\gamma_{ij}^d)$ for  $i= 1, \ldots, C$,  $j= 0 \text{ and } 1$\\
Therefore, for a document $d$, 
\begin{align*}
q(\Theta^d) = \prod_{i=1}^C q(\lambda^d) \prod_{i=1}^C \prod_{j=0}^1 q(\theta_{ij}^d) \prod_{n=1}^N\prod_{i=1}^C q(u_{ni}^d) q(z_{ni}^d) 
\end{align*}
The E-step involves computing the document-specific variational parameters $\{ \delta^d, \Delta^d, \gamma^d, \phi^d \}$ assuming a fixed value for the parameters $\Pi = \{ \alpha, \xi, \rho, \beta \}$. As a consequence of the mean field assumptions on the variational distributions, we get the following update rules for the distributions. From now on, when clear from context we omit the superscript $d$.
\begin{align}
\log q(z)&= \mathbb{E}_{\Theta \setminus z} \left[ p(\bold{d}, \Theta) \right] \propto \mathbb{E}_{u,\lambda,\theta} \left[
\log p(z|u, \lambda, \theta) \right] + \log p(w|z, \beta) \nonumber \\ &\propto \sum_{n=1}^N\sum_{t=1}^T  z_{nt} \left[\sum_{i=1}^C \sum_{j=0}^1 \mathbb{E}[u_{ni}] \mathbb{E}[\lambda_{ij}] \mathbb{E}[\log \theta_{ijt}]
+ \sum_{j=1}^V w_{nj} \log \beta_{tj}\right]
\label{eqn:z-var-deriv}
\end{align}
In the computation of the expectation of   $\mathbb{E}_{\Theta \setminus z} \left[ p(\bold{d}, \Theta) \right]$ in Eqn \ref{eqn:z-var-deriv}, the terms in $p(\bold{d}, \Theta)$ that are a function of $z$ need to be considered as the rest of the terms contribute to the normalizing constant for the density function $q(z)$. Hence expectations of $\log p(z|u, \lambda, \theta)$ (Eqn \ref{eqn:z-pdf}) and $\log p(w|z,\beta)$ (Eqn \ref{eqn:w-pdf}) must be taken with respect to $u,  \lambda, \theta$.
Therefore ,
\begin{align}
\log \phi_{nt} &\propto  \sum_{i=1}^C \sum_{j=0}^1 \mathbb{E}[u_{ni}] \mathbb{E}[\lambda_{ij}] \mathbb{E}[\log \theta_{ijt}]
+ \sum_{j=1}^V w_{nj} \log \beta_{tj} \nonumber\\ &= 
\sum_{i=1}^C \sum_{j=0}^1 \delta_{ni} \Delta_i^j (1-\Delta_i)^{1-j} \mathbb{E}[\log \theta_{ijt}]
+ \sum_{j=1}^V w_{nj} \log \beta_{tj}
\end{align}
Similarly, the updates for the other variational parameters follows.
\begin{align}
\label{eqn:u-var-deriv}
\log q(u) &= \mathbb{E}_{\Theta \setminus u} \left[ p(\bold{d}, \Theta) \right] = \mathbb{E}_{\Theta \setminus u} \left[ \log p(u) + p(z| u, \lambda, \theta) \right] \nonumber \\ &\propto \sum_{n=1}^N \sum_{i=1}^C u_{ni} \log 1/C + \sum_{n=1}^N \sum_{t=1}^T \sum_{i=1}^C \sum_{j=0}^1 u_{ni} \mathbb{E}[\lambda_{ij}] \mathbb{E}[z_{nt}] \mathbb{E}[\log \theta_{ijt}] 
\end{align}
Therefore,
\begin{align}
\log \delta_{ni} \propto \log\frac{1}{C} + \sum_{t=1}^T \phi_{nt}\Delta_i \mathbb{E}\left[ \log \theta_{i1t}\right]+
\phi_{nt}(1-\Delta_i )\mathbb{E}\left[ \log \theta_{i0t}\right]
\end{align}

\begin{align}
\label{eqn:theta-var-deriv}
\log &\; q(\theta)  = \mathbb{E}_{\Theta \setminus \theta} \left[ p(\bold{d}, \Theta) \right] \propto \mathbb{E} \left[ p(\theta|\alpha) + p(z|u, \lambda, \theta) \right] \\
&= \sum_{i=1}^C \sum_{j=0}^1 \sum_{t=1}^T (\alpha_{ijt}-1) \log \theta_{ijt} + \sum_{n=1}^N \sum_{i=1}^C \sum_{j=0}^1 \sum_{t=1}^T \mathbb{E}[u_{ni}] E[z_{nt}] E[\lambda_{ij}] \log  \theta_{ijt} \\&=\sum_{i=1}^C \sum_{j=0}^1 \sum_{t=1}^T (\alpha_{ijt}-1) \log \theta_{ijt} + \sum_{n=1}^N \sum_{i=1}^C \sum_{j=0}^1 \sum_{t=1}^T \delta_{ni} \phi_{nt} \Delta_i^j (1-\Delta_i)^{1-j} \log  \theta_{ijt} \nonumber \\ &= \sum_{i=1}^C \sum_{j=0}^1 \sum_{t=1}^T (\gamma_{ijt}- 1) \log \theta_{ijt}
\end{align}
where,
\begin{align}
\gamma_{ijt} = \alpha_{ijt} + (\Delta_i)^j (1-\Delta_i)^{1-j}
\sum_{n=1}^N \delta_{ni} \phi_{nt} 
\end{align}
\begin{align}
\log q(\lambda)  &= \mathbb{E}_{\Theta \setminus \lambda} \left[ p(\bold{d}, \Theta) \right] \propto \mathbb{E} \left[ \log p(\lambda|\xi) + \log p(z|u, \lambda, \theta) + \log p(y|\lambda, \rho) \right] \nonumber \\ &= \sum_{i=1}^C  \lambda_i \log \xi_i +  (1-\lambda_i) \log (1-\xi_i) \nonumber\\&\;\;\; +  \sum_{n=1}^{N} \sum_{t=1}^T\sum_{i=1}^C\sum_{j=0}^{1}\lambda_{i}^j (1-\lambda_i)^{1-j} \mathbb{E}[u_{ni}]  \mathbb{E}[z_{nt}] \mathbb{E}[\log \theta_{ijt}] \nonumber \\ &\;\;\;+ \sum_{j=1}^{K}\sum_{i=1}^C \left[\lambda_i y_{ij} + (1-\lambda_i)(1 - y_{ij})\right] \log \rho_j  \nonumber \\&\;\;\;\;\;\;\; + \left[ (1- \lambda_i) y_{ij} + \lambda_i(1 - y_{ij})\right] \log  1 - \rho_j
\end{align}

\subsection{E-step Updates for ML-PA-LDA-C}
In the E-step, the following steps need to be performed iteratively for each document $d$.
\begin{align}
\label{eqn:e-step-Delta}
\log \Delta_i^d &\propto \log \xi_i + \sum_{j=1}^K y_{ij}^d \log \rho_j + (1-y_{ij}^d)\log 1-\rho_j + \sum_{n=1}^{N_d} \sum_{t=1}^T \delta_{ni}^d \phi_{nt}^d \mathbb{E}\left[\log \theta_{i1t}^d \right]
\end{align}
\begin{align}
\label{eqn:e-step-1-Delta}
\log (1-\Delta_i^d) &\propto \log 1-\xi_i + \sum_{j=1}^K (1-y_{ij}^d) \log \rho_j + y_{ij}^d\log (1-\rho_j ) + \sum_{n=1}^{N_d} \sum_{t=1}^T \delta_{ni}^d \phi_{nt}^d \mathbb{E}\left[ \log \theta_{i0t}^d \right] 
\end{align}
\begin{align}
\label{egn:e-step-phi}
\log \phi_{nt}^d \propto \sum_{i=1}^C \delta_{ni}^d \left[ \Delta_{i}^d\mathbb{E}\left[ \log \theta_{i1t}^d\right] + (1-\Delta_{i}^d)\mathbb{E}\left[ \log \theta_{i0t}^d\right]\right] 
+ \sum_{j=1}^V w_{nj}^d \log \beta_{tj}
\end{align}
\begin{align}
\label{egn:e-step-delta}
\log \delta_{ni}^d \propto \log\frac{1}{C} + \sum_{t=1}^T \phi_{nt}\Delta_i^d \mathbb{E}\left[ \log \theta_{i1t}^d\right]+
\phi_{nt}^d(1-\Delta_i^d )\mathbb{E}\left[ \log \theta_{i0t}^d\right]
\end{align}
\begin{align}
\label{egn:e-step-gamma}
\gamma_{ijt}^d = \alpha_{ijt} + (\Delta_i^d)^j (1-\Delta_i^d)^{1-j}
\sum_{n=1}^N \delta_{ni}^d \phi_{nt}^d 
\end{align}
In all the above update rules,
$\mathbb{E}[\log \theta_{ijt}^d] = \psi(\gamma_{ijt}) - \psi(\sum_{t'=1}^T \gamma_{ijt'}^d)$, where $\psi(.)$ is the digamma function.
For the non-crowd model ML-PA-LDA, the variational parameter $\Delta_i$ is absent as $\lambda_i$ is observed. Therefore the E-step boils down to computing the updates for $\phi, \delta $ and $\gamma$ from Eqns \ref{egn:e-step-phi}, \ref{egn:e-step-delta} and \ref{egn:e-step-gamma} with  $\Delta_i$ replaced by $\lambda_i$.
\subsection{M-step Updates for ML-PA-LDA-C}
In the M-step, the parameters $\xi$, $\rho$, $\beta$ and $\alpha$ are estimated using the values of $\Delta^d, \phi_{nt}^d, \delta_{ni}^d, \gamma_{ijt}^d$ estimated from the E-step. The function $L(\Theta)$ in Eqn \ref{eqn:elbo} is maximized with respect to the parameters $\pi$ yielding the following update equations.\\
\textbf{Updates for $\xi$:}\\

\begin{align}
\xi_i = \frac{\sum_{d=1}^D \Delta_i^d}{D} \;\;\; \text{ for }  i= 1, \ldots, C.
\label{eqn:m-step-xi}
\end{align}
Intuitively,  Eqn \ref{eqn:m-step-xi} makes sense as $\xi_i$ is the probability that any document in the corpus belongs to class $i$. $\Delta_i^d$ is the probability that document $d$ belongs to class $i$ and is computed in the E-step. Therefore $\xi_i$ is an average of $\Delta_i^d$ over all documents.\\
\textbf{Updates for $\rho$:}\\
for $j = 1, \ldots, K$:
\begin{align}
\label{eqn:m-step-rho}
\displaystyle
\rho_j = \frac{\sum\limits_{d=1}^D \sum\limits_{i=1}^C \mathbbm{1}\left[y_{ij}^d \neq -1\right] \left[y_{ij}^d \Delta_i^d + (1- y_{ij}^d)(1- \Delta_i^d)\right]}{\sum\limits_{d=1}^D \sum\limits_{i=1}^C \mathbbm{1}\left[y_{ij}^d \neq -1\right]\left[y_{ij}^d \Delta_i^d + (1- y_{ij}^d)(1- \Delta_i^d) + y_{ij}^d (1-\Delta_i^d) + (1-y_{ij}^d)\Delta_i^d\right]}
\end{align}
From Eqn \ref{eqn:m-step-rho}, we observe that $\rho_j$ is the fraction of times that crowd-worker $j$ has provided a label that is consistent with the probability estimate $\Delta^d_i$ over all classes $i$. The implicit assumption is that every crowd-worker has provided at least one label, otherwise such a crowd-worker need not be considered in the model.\\
\textbf{Updates for $\beta$}: \\
for $t=1, \ldots, T$; for $j=1, \ldots, V$:
\begin{align}
\label{m-step-beta}
\beta_{tj} = \frac{\sum_{d=1}^D \sum_{n=1}^{N_d} w_{nj}^d \phi_{nt}^d}{\sum_{d=1}^D N_d}
\end{align}
Intuitively, the variational parameter $\phi_{nt}^d$ is the probability that the word $w_{n}^d$ is associated with topic $t$. Having updated this parameter in the E-step, $\beta_{tj}$ computes the fraction of times the word $j$ is associated with topic $t$ by giving a weight $\phi_{nt}^d$ to its occurrence in document $d$.\\
\textbf{Updates for $\alpha$}: \\
There do not exist closed form updates for $\alpha$ parameters. Hence we use Newton Raphson method to iteratively obtain the solution as follows.
\begin{align}
\alpha_{ijr}^{t+1} = \alpha_{ijr}^t - \frac{g_r - c}{h_r}
\end{align}
where, $c= \frac{\sum_{\tau = 1}^T g_\tau/h_\tau}{z^{-1} + \sum_{\tau = 1}^T 1/h_\tau}$, 
$z = D\psi'\left( \sum_{t'=1}^T \alpha_{ijt'}^t \right)$, $h_\tau = -D\psi'(\alpha_{ijr}^t)$, \\ $g_r = D\left[\psi\left( \sum_{\tau=1}^T \alpha_{ij\tau}^t \right) - \psi\left( \alpha_{ijr}^t\right)  \right] + \sum_{d=1}^D \left[ \psi\left(\gamma_{ijr}^d\right) - \psi\left(\sum_{tau=1}^T\gamma_{ij\tau}^d\right) \right]$ 

The M-step updates for $\beta$ and $\alpha$ involved in ML-PA-LDA (non-crowd version) are same as the updates in the crowd version, ML-PA-LDA-C.  The parameter $\rho$ is absent in ML-PA-LDA.
Eqn \ref{eqn:m-step-xi}, with $\Delta_i^d$ replaced by $\lambda_i^d$ (as in the E-step) is used to update $\xi_i$.
\section{Smoothing}
\begin{figure}
\centering
\includegraphics[scale=0.55]{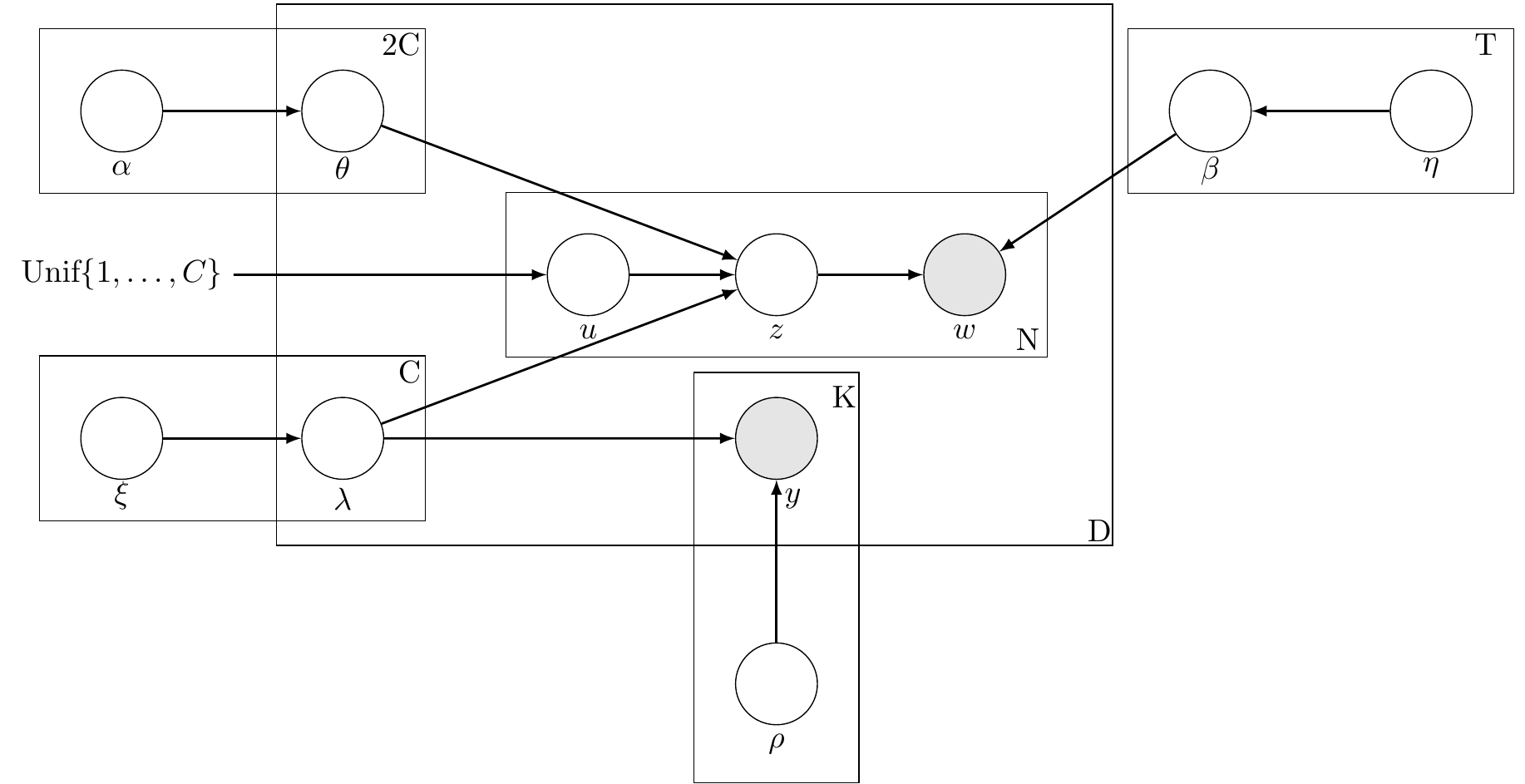}
\caption{Graphical model representation for multi-label classification from the crowd when smoothing parameters are included.}
\label{fig:graph-model-smooth}
\end{figure}

In the model described in Section \ref{sec:multi-label-model}, we modeled $\beta$ to be a parameter that governs the multinomial distributions for generating the words from each topic. In general, a new document can include words that have not been encountered in any of the training documents. The unsmoothed model described earlier does not handle this issue. In order to handle this, we must ``smoothen'' the multinomial parameters involved. One way to perform smoothing is to model $\beta$ as a multinomial random variable with parameters $\eta$. The corresponding graphical model is provided in \Cref{fig:graph-model-smooth}. Again due to the intractable nature of the computations, we model the variational distribution for $\beta$ as $\beta \sim \text{Mult}(\chi)$. 
We estimate variational parameter $\chi$ in the E-step of variational EM using Eqn \ref{eqn:e-step-chi} assuming $\eta$ is known.
\begin{align}
\label{eqn:e-step-chi}
\chi_{tj} = \eta_{tj} + \sum_{d=1}^D \sum_{n=1}^{N_d} \phi_{nt}^d w_{nj}^d
\end{align}
The model parameter $\eta$ is estimated in the M-step using Newton Raphson method as follows.
\begin{align}
\eta_{ir}^{t+1} = \eta_{ir}^t - \frac{g_r - c}{h_r}
\end{align}
where, $c= \frac{\sum_{\tau = 1}^V g_\tau/h_\tau}{z^{-1} + \sum_{\tau = 1}^T 1/h_\tau}$, 
$z = \psi'\left( \sum_{j'=1}^V \eta_{ij'}^t \right)$, $h_r = -\psi'(\eta_{ir}^t)$, \\ $g_r = \left[\psi\left( \sum_{j'=1}^V \eta_{ij'}^t \right) - \psi\left( \eta_{ir}^t\right)  \right] + \left[ \psi\left(\chi_{ir}\right) - \psi\left(\sum_{j'=1}^V\chi_{ij'}\right) \right]$. 
 The steps for the derivation are similar to the steps for non-smooth version.
\section{Experiments}
In order to test the efficacy of the proposed techniques, we evaluate our model on datasets from several domains.
\subsection{Dataset Descriptions}
We have carried out our experiments on several datasets from the text domain as well as non-text domain. We now describe the datasets and the preprocessing steps below.
\subsubsection{\underline{Text Datasets:}}
In the text domain, we have performed studies on the Reuters-21578, Bibtex and Enron datasets.\\
\textbf{Reuters-21578:}
The Reuters-21578 dataset \cite{Lichman2013} is a collection of documents with news articles. The original corpus had 10,369 documents and a vocabulary of 29930 words. We performed stemming using the Porter Stemmer algorithm \cite{porter1980algorithm} and also removed the stop words.  From this set the words which occurred more than 50 times across the corpus were retained and only documents which contained more than 20 words were retained.  Finally the most commonly occurring top 10 labels were retained namely acq, crude, earn, fx, grain , interest, money, ship, trade, wheat. This led to a total of 6547 documents and a vocabulary of size 1996. Of these, a random 80\% was used as training set and the remaining 20\% as test. \\
\textbf{Bibtex:} The Bibtex dataset \cite{bibtexDataset} was released as part of the ECML-PKDD 2008 Discovery Challenge. The task is to assign tags such as physics, graph, electrochemistry etc to bibtex entries. There are a total of 4880 and 2515 entries in the training set and test respectively. The size of the vocabulary is 1836 and the number of tags is 159.\\
\textbf{Enron: } The Enron dataset \cite{mulan} is a collection of emails for which a set of pre-defined categories are to be assigned. There are a total of 1123 and 573 training and test instances respectively with a vocabulary of 1001 words. The total number of email tags are 53.
\subsubsection{\underline{Non-text Datasets:}}
We also evaluate our model on datasets from domains other than text, where the notion of words is not explicit. \\
\textit{Converting real valued features to words:} Since we assume a bag-of-words model, we must replace every real-valued feature with a `word' from a `vocabulary'. We begin by choosing an appropriate size for the vocabulary. Thereafter, we collect every real number which occurs across features and instances in the corpus into a set. We then cluster this set into $V$ clusters, using the k-means algorithm, where $V$  is the size of the vocabulary previously chosen. Therefore, each real valued feature has a new representative word given by the nearest cluster center to the feature under consideration. The corpus is then generated according to this new feature representation scheme.\\
\textbf{Yeast:} The Yeast dataset \cite{ranksvm} contains a set of genes which may be associated with several functional classes. There are 1500 training examples and 917 examples in the test set with a total of 14 classes and 103 real valued features.\\
\textbf{Scene:} The Scene dataset \cite{boutell2004learning} is a dataset of images. The task is to classify images into the following 6 categories- beach, sunset, fall, field, mountain, urban. The dataset contains 1211 instances in the training set and 1196 instances in the test set with a total of 294 real valued features.

In our experiments, we use the measures, accuracy across classes,  micro-f1 score and average class log likelihood on the test sets to evaluate our model. 

The average class log-likelihood on the test instances is computed as follows:
\begin{align*}
\text{log-l}  =\frac{ \sum_{d=1}^{D_{test}}\sum_{i=1}^C 
\lambda_i^d \log \Delta_i^d + (1-\lambda_i^d) \log (1-\Delta_i^d)} {D_{test} \times C}
\end{align*}
where $D_{test}$ is the number of instances in the test set. The details of computation of the other measures can be found in the survey \cite{Gibaja2015}.
\subsection{Results: ML-PA-LDA (Non-Crowd Version)}
We run our model first assuming labels from a perfect source.
\newcolumntype{P}[1]{>{\centering\arraybackslash}p{#1}}

In \Cref{tab:ml-pa-lda-performance} we compare the performance of our non-annotator model vs other methods such as RAKel, Monte Carlo Classifier Chains (MCC) \cite{ReadICASSP}, Binary Relevance Method - Random Subspace (BRq) \cite{Read2011}, Bayesian Chain Classifiers (BCC) \cite{Zaragoza2011} and SLDA. 
BCC \cite{Zaragoza2011} is a probabilistic method which constructs a chain of classifiers by modeling the dependencies between the classes using a bayesian network. MCC instead uses a monte-carlo strategy to learn the dependencies. BRq improves upon binary relevance methods of combining classifiers by constructing an ensemble. As mentioned earlier RAKel draws subsets of the classes, each of size $k$ and constructs ensemble classifiers.
The implementations of RAKel, MCC, BRq and BCC provided by  Meka (http://meka.sourceforge.net/) were used. For SLDA the code provided by the authors was used.
On the reuters dataset, ML-PA-LDA (without the annotators) performs significantly better than SLDA. In fact, ML-PA-LDA-C trained with noisy labels provided by simulated annotators also performs much better than SLDA. On the bibtex and enron datasets, ML-PA-LDA does better than SLDA while ML-PA-LDA-C is not far. On scene and yeast datasets, ML-PA-LDA, ML-PA-LDA-C and SLDA give the same performance. It is to be noted that these datasets, known to be hard, are from the images and biology domains respectively. As can be seen from the table, our model gives a better overall performance than SLDA and also does not require training $C$ binary classifiers. This advantage is a significant one, especially in datasets such as bibtex where the number of classes is 159. 

\begin{figure}[h!]
\centering
\begin{minipage}{0.49\textwidth}
\includegraphics[scale=0.3]{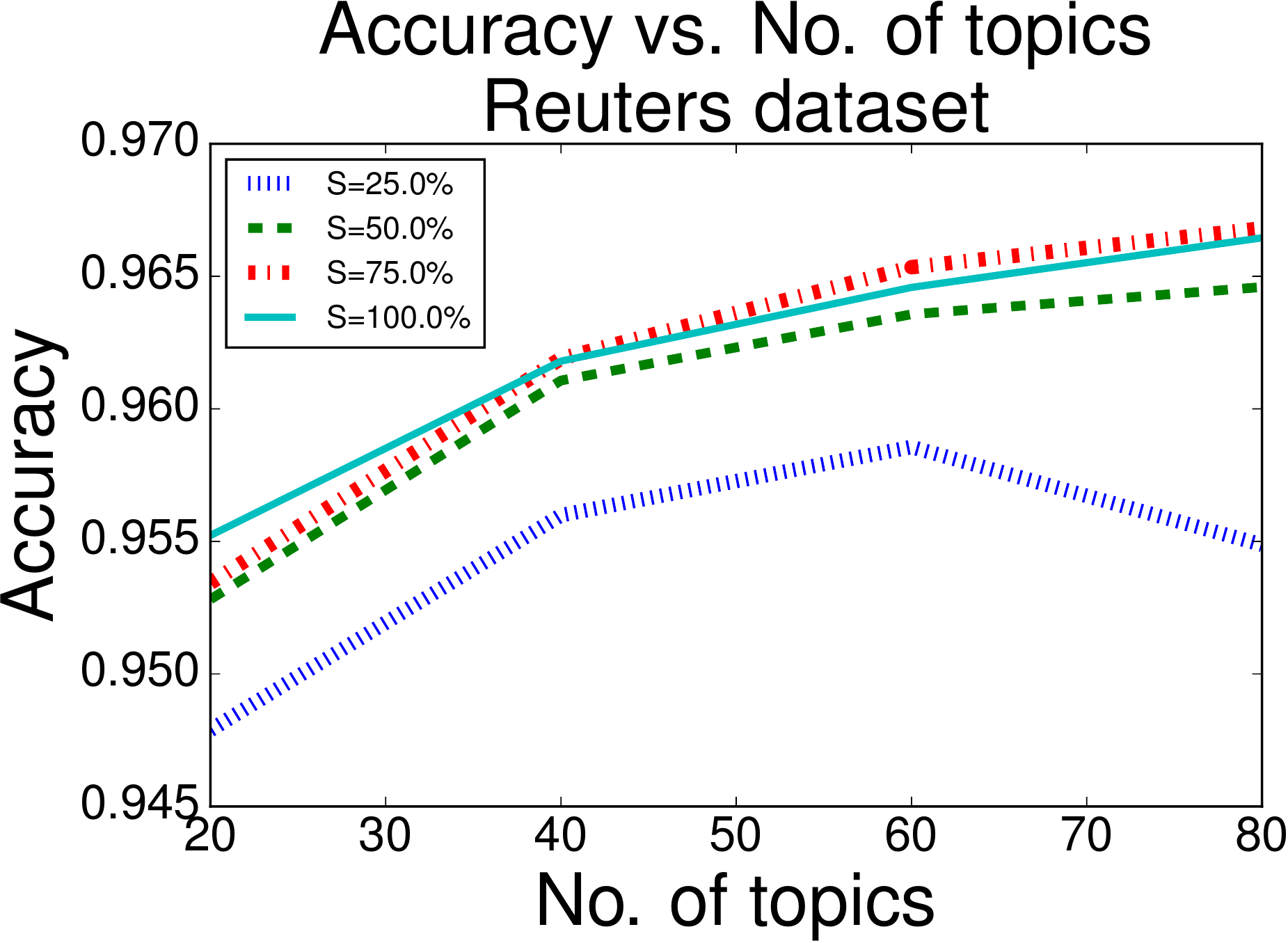}
\subcaption{}
\end{minipage}
\begin{minipage}{0.49\textwidth}
\includegraphics[scale=0.3]{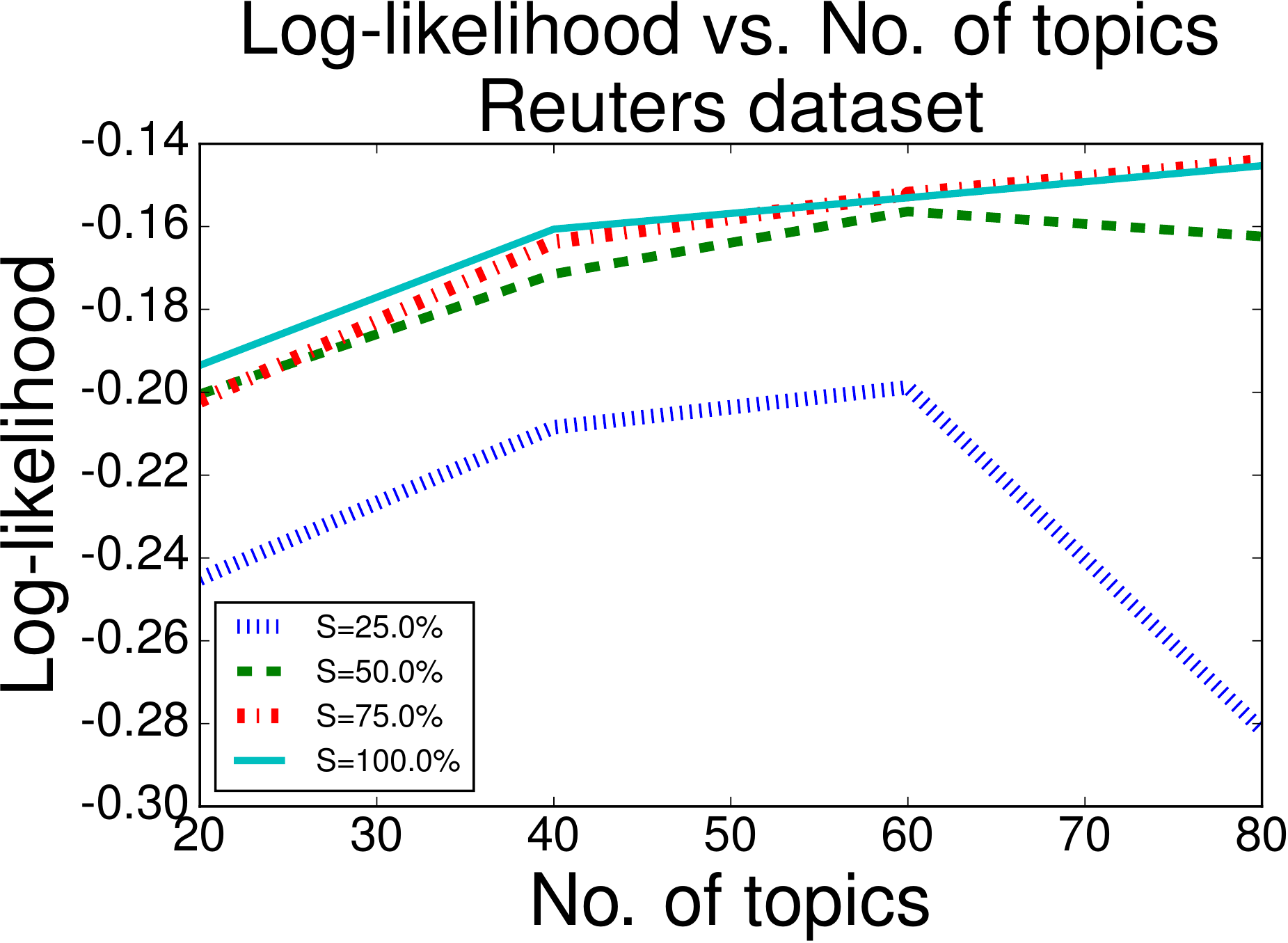}
\subcaption{}
\end{minipage}

\begin{minipage}{1.0\textwidth}
\centering
\includegraphics[scale=0.3]{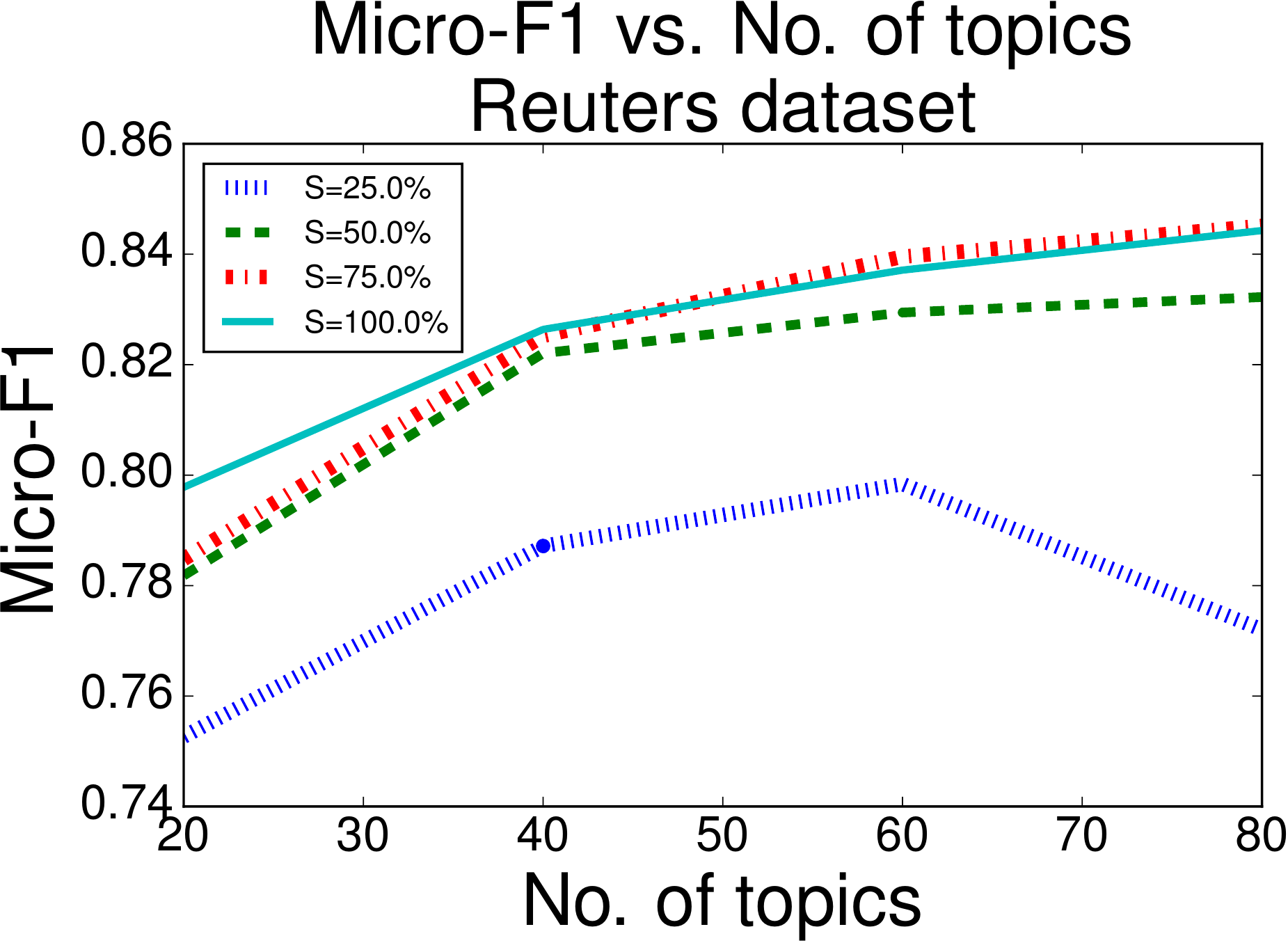}
\subcaption{}
\end{minipage}
\caption{Performance of ML-PA-LDA (non-crowd version) on Reuters dataset. S is the fraction of the dataset used for training the model. Other datasets (omitted for space) follow a similar trend.}
\label{fig:non-ann-topics-dataset-size}
\end{figure}

\begin{table}[h!]
\centering
\begin{tabular}[c]{|P{0.1\textwidth}|P{0.1\textwidth}|P{0.1\textwidth}|P{0.1\textwidth}|P{0.1\textwidth}|P{0.1\textwidth}|P{0.1\textwidth}|P{0.1\textwidth}|}\hline
Dataset & RAKel (J48) & MCC & BRq & BCC & SLDA & ML-PA-LDA & ML-PA-LDA-C  \\\hline
reuters &  0.881 & 0.876 & 0.863 & 0.867 & 0.897 &  \textbf{0.969}  & \textbf{0.9426}\\\hline
bibtex & 0.293  & 0.290 & 0.309 & 0.299 & 0.984 & \textbf{0.984} & \textbf{0.981}\\\hline
enron & 0.402 & 0.389  & 0.430 & 0.411 & 0.937 & \textbf{0.939} &
\textbf{0.938} \\\hline 
scene & 0.577 & 0.580 & 0.550 & 0.594 & \textbf{0.823} & \textbf{0.823} & \textbf{0.818} \\\hline
yeast & 0.415 &  0.432 & 0.462 & 0.413 & \textbf{0.767} & \textbf{0.767} & \textbf{0.767} \\\hline
\end{tabular}
\caption{Comparison of average accuracy of various multilabel classification techniques}
\label{tab:ml-pa-lda-performance}
\end{table}
We compared the performance of our algorithm with the size of the datasets used for training as well as the number of topics used. The results of our model are shown in \Cref{fig:non-ann-topics-dataset-size}. An increase in the size of the dataset improves the performance of our model with respect to all the measures in use. Similarly an increase in the number of topics generally improves the measures under consideration. A striking observation is the low accuracy, log likelihood and micro-f1 scores associated with the model when the number of topics = 80 (eight times the number of classes) and the size of the dataset is low (S=25\%). This is expected as the number of parameters to be estimated is too large to be learned using very few training examples. However as more training data is available, the model achieves enhanced performance. This observation is consistent with Occam's razor \cite{OccamRazor}.
\subsection{Results: ML-PA-LDA-C (Crowd Version) }
To verify the performance of the annotator model where the labels are provided by multiple noisy annotators, we simulated 50 annotators with varying qualities. The $\rho$ values of the annotators were sampled from a uniform distribution. For 10 of these annotators, $\rho$ was sampled from $U[0.51, 0.65]$. For another 20 of them, $\rho$ was sampled from $U[0.66, 0.85]$
and for the remaining 20 of them $\rho$ was sampled from $U[0.86, 0.9999]$. This captures the heterogenity in the annotator qualities. For each document in the training set, a random $10\% \;(= 5)$ annotators were picked for generating the noisy labels. 

In \Cref{tab:ann-non-ann-comparison} we compare the performance of the annotator model against our non-annotator model version. We find that the performance of ML-PA-LDA-C is close to that of ML-PA-LDA and most often better than SLDA, inspite of having access to only noisy labels. We also report `Ann RMSE' which is the L2 norm  of the difference in predicted qualities of the annotators vs the true qualities. 
\begin{align}
\text{Ann RMSE} = \sqrt{\sum_{j=1}^K | \hat{\rho}_j - \rho_j |^2 / K}
\end{align}
where  $\hat{\rho}_j$ is the quality of annotator $j$ as predicted by our variational EM algorithm and $\rho_j$ is the true, hidden annnotator quality. We find that `Ann RMSE' decreases as more training data is available showing the efficacy of our model for learning the qualities of the annotators.

\begin{table}[h!]
\centering
\begin{tabular}[c]{|P{0.1\textwidth}|P{0.16\textwidth}|P{0.16\textwidth}||P{0.18\textwidth}| P{0.18\textwidth}|| P{0.15\textwidth}| }
\hline
\parbox[c]{6cm}{\% of \\training \\set used }& \parbox[c]{6cm}{ ML-PA-LDA\\ avg accuracy } & \parbox[c]{6cm}{ML-PA-LDA\\ avg  microf1 } & \parbox[c]{6cm}{ML-PA-LDA-C \\ avg accuracy} & \parbox[c]{5cm}{ML-PA-LDA-C\\avg microf1 } &  \parbox[c]{6cm}{Ann RMSE } \\\hline %
10	&  0.949	&	0.762 & 0.927 	 & 	 0.616 & 0.023 	 \\
30	&  0.953	& 0.784 & 0.930 	 & 0.619  & 0.014\\
50	&  0.955    &	 0.787 & 0.936 & 0.629 & 0.011 \\
70	& 0.961 	& 0.828 &0.937 	&	 0.650 & 0.010 \\
100	&  0.969 & 0.829 & 0.942 & 0.669 & 0.009\\ \hline
\end{tabular}
\caption{Comparison of performances of non-annotator model and annotator model for a fixed number of topics (= 20)}
\label{tab:ann-non-ann-comparison}
\end{table}

\begin{figure}[h!]
\centering
\begin{minipage}{0.49\textwidth}
\includegraphics[scale=0.3]{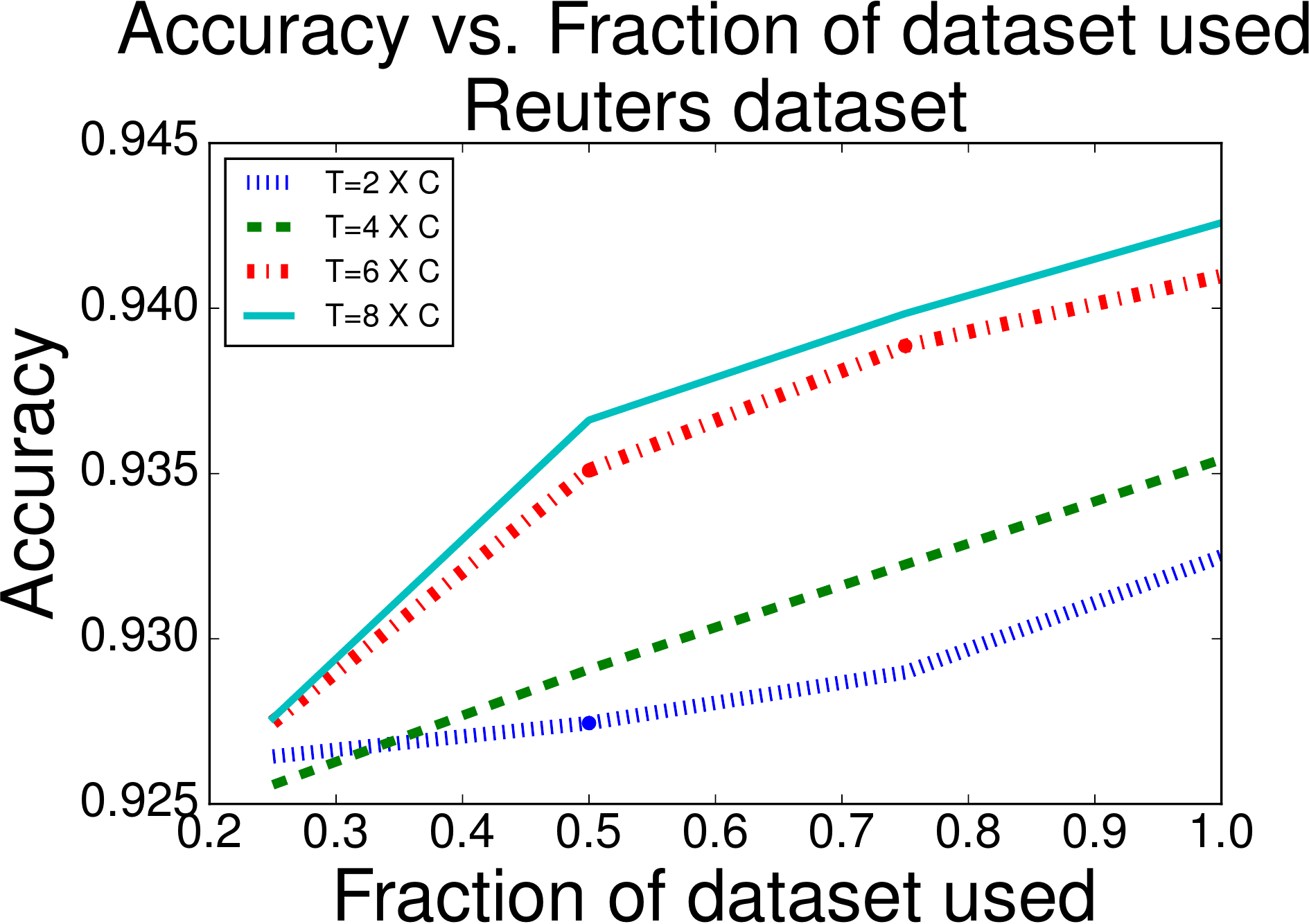}
\subcaption{}
\end{minipage}
\begin{minipage}{0.49\textwidth}
\includegraphics[scale=0.3]{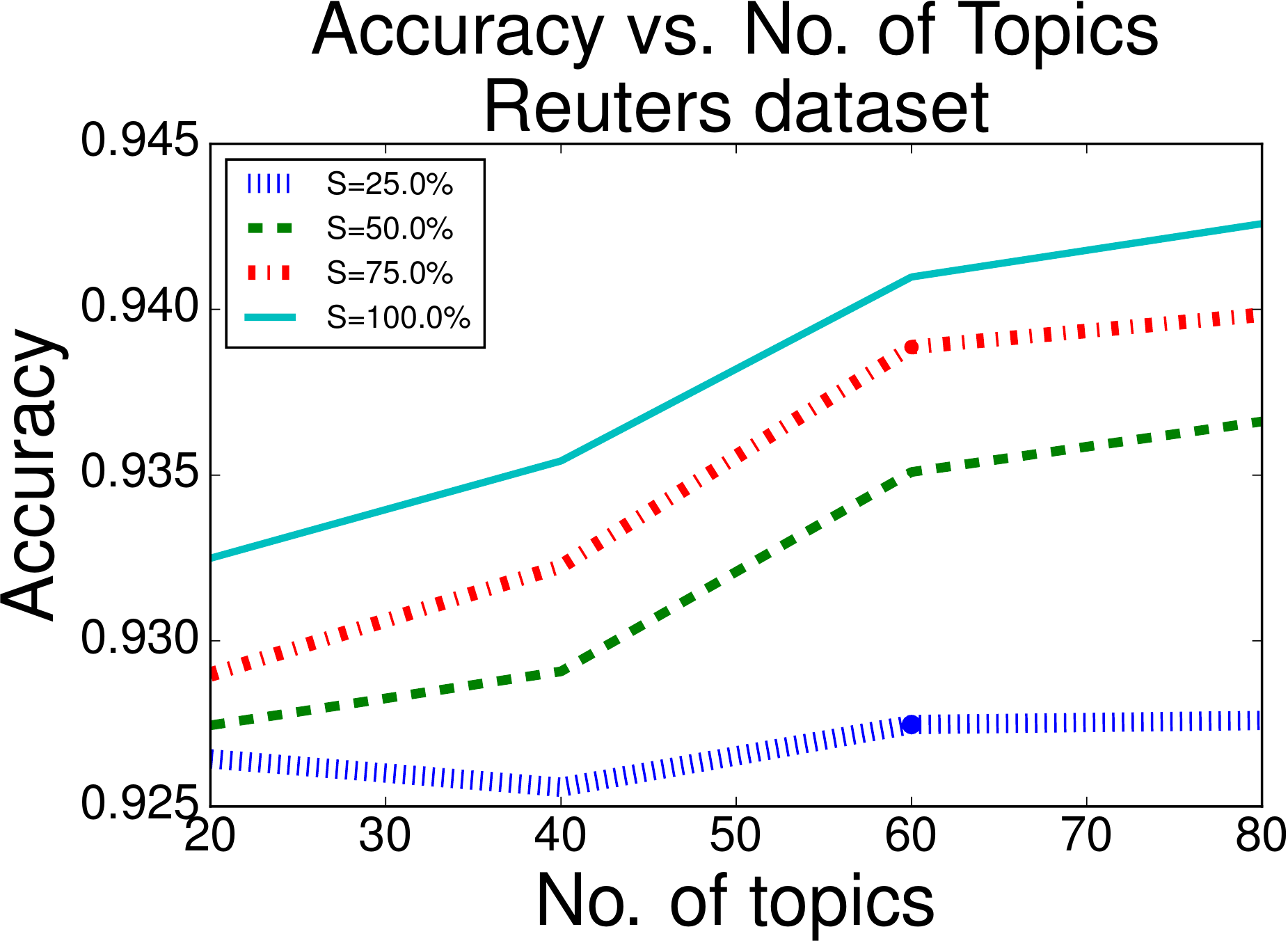}
\subcaption{}
\end{minipage}

\begin{minipage}{0.49\textwidth}
\centering
\includegraphics[scale=0.3]{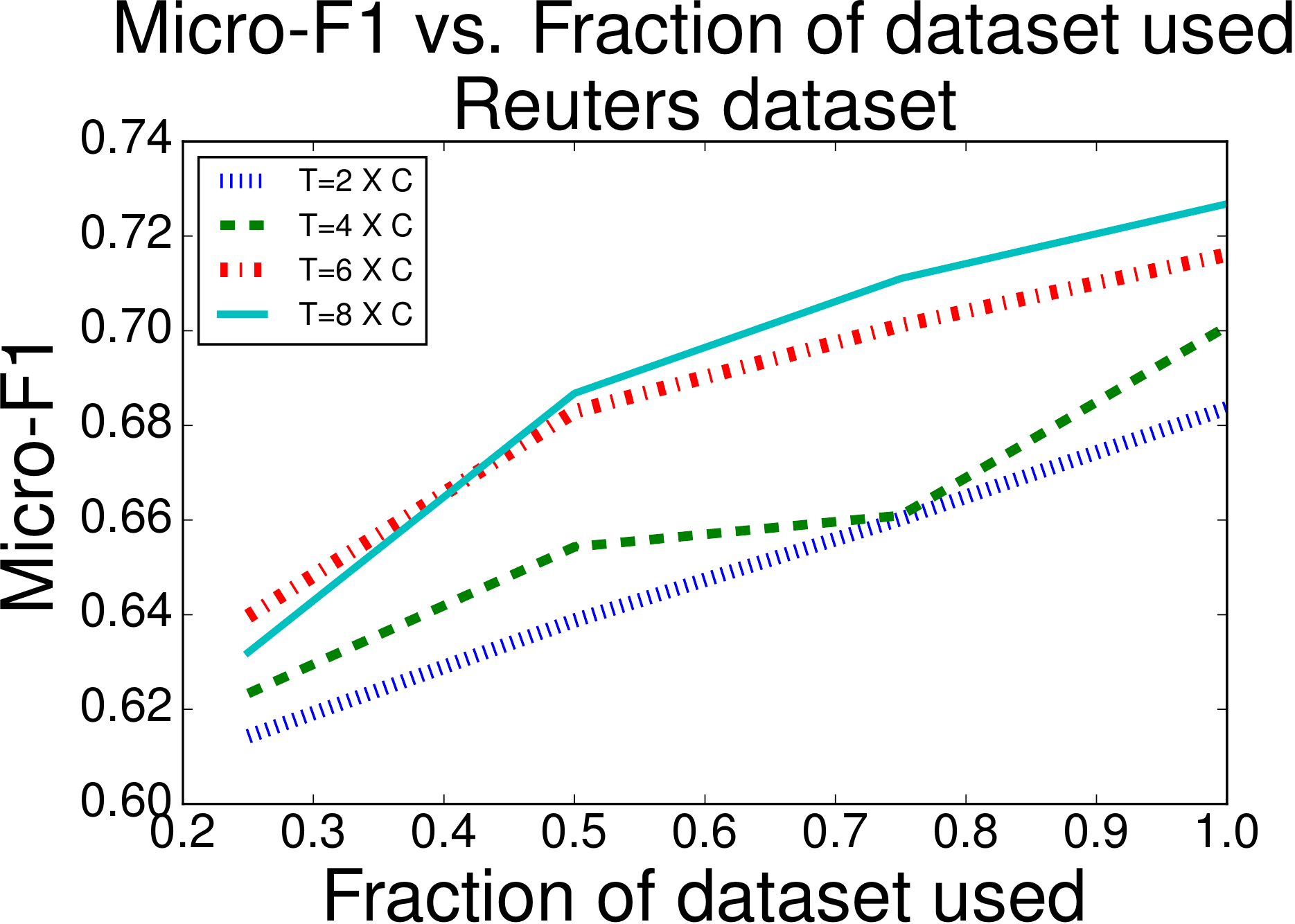}
\subcaption{}
\end{minipage}
\begin{minipage}{0.49\textwidth}
\includegraphics[scale=0.3]{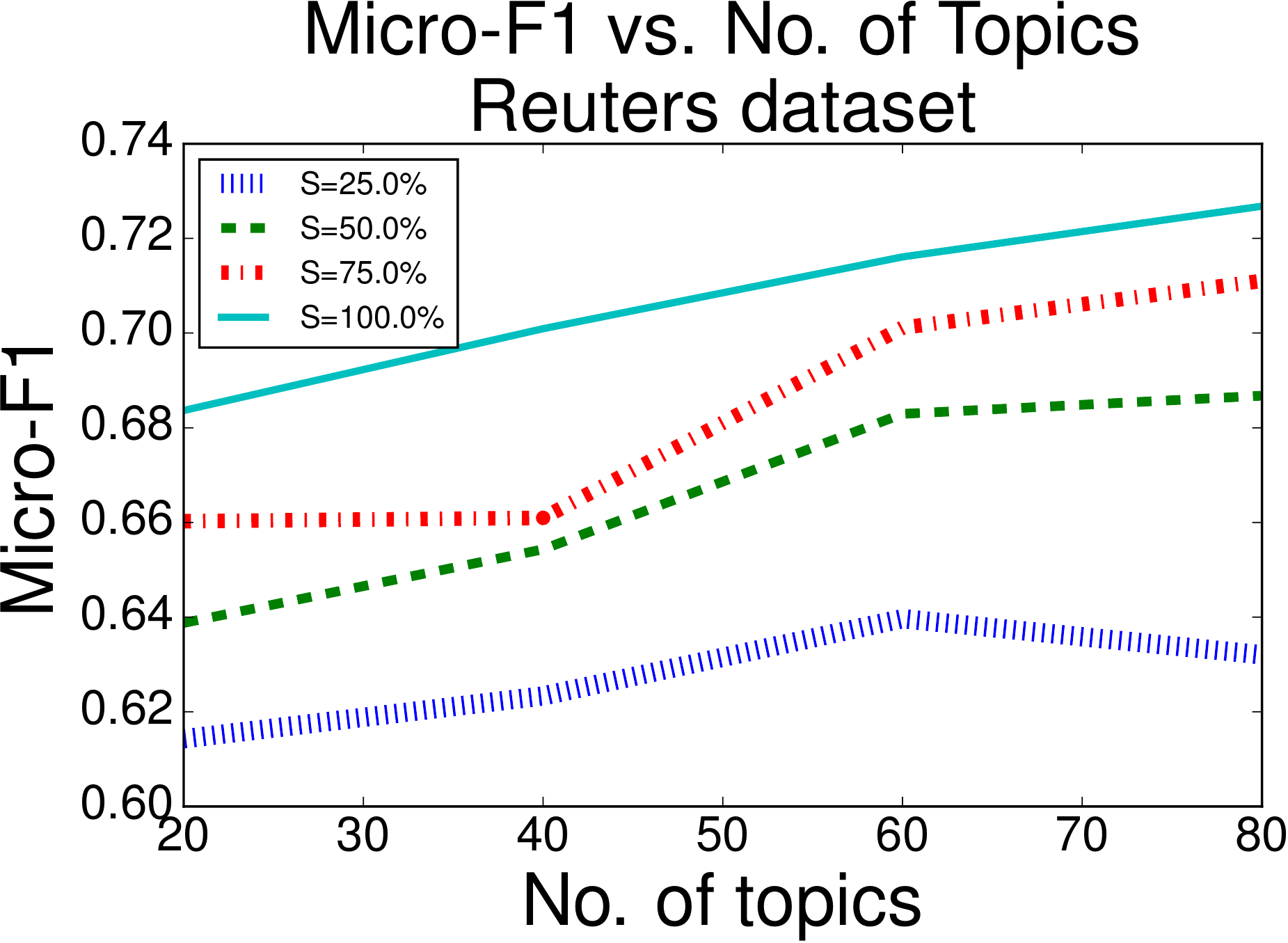}
\subcaption{}
\end{minipage}

\begin{minipage}{0.49\textwidth}
\includegraphics[scale=0.3]{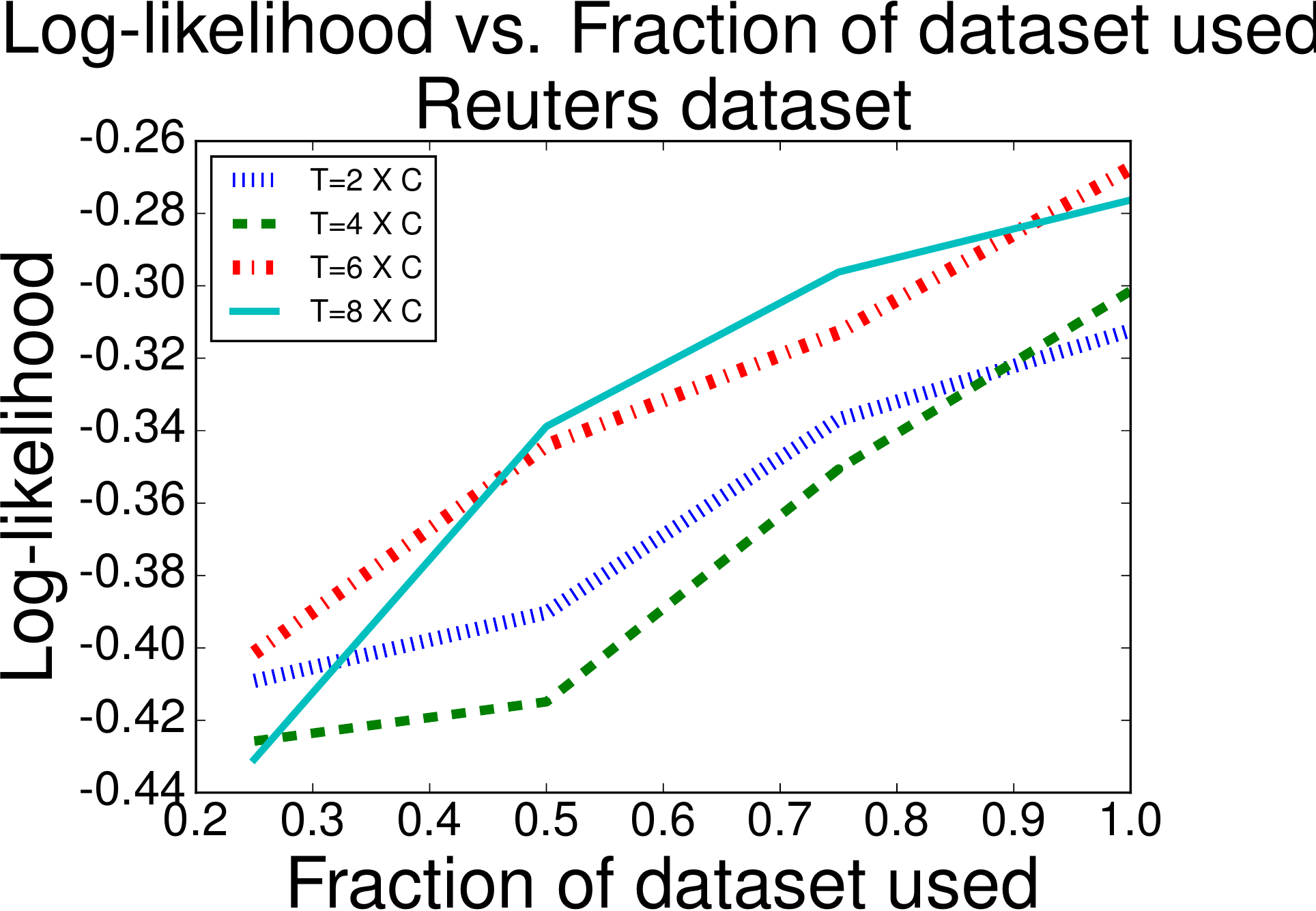}
\subcaption{}
\end{minipage}
\begin{minipage}{0.49\textwidth}
\centering
\includegraphics[scale=0.3]{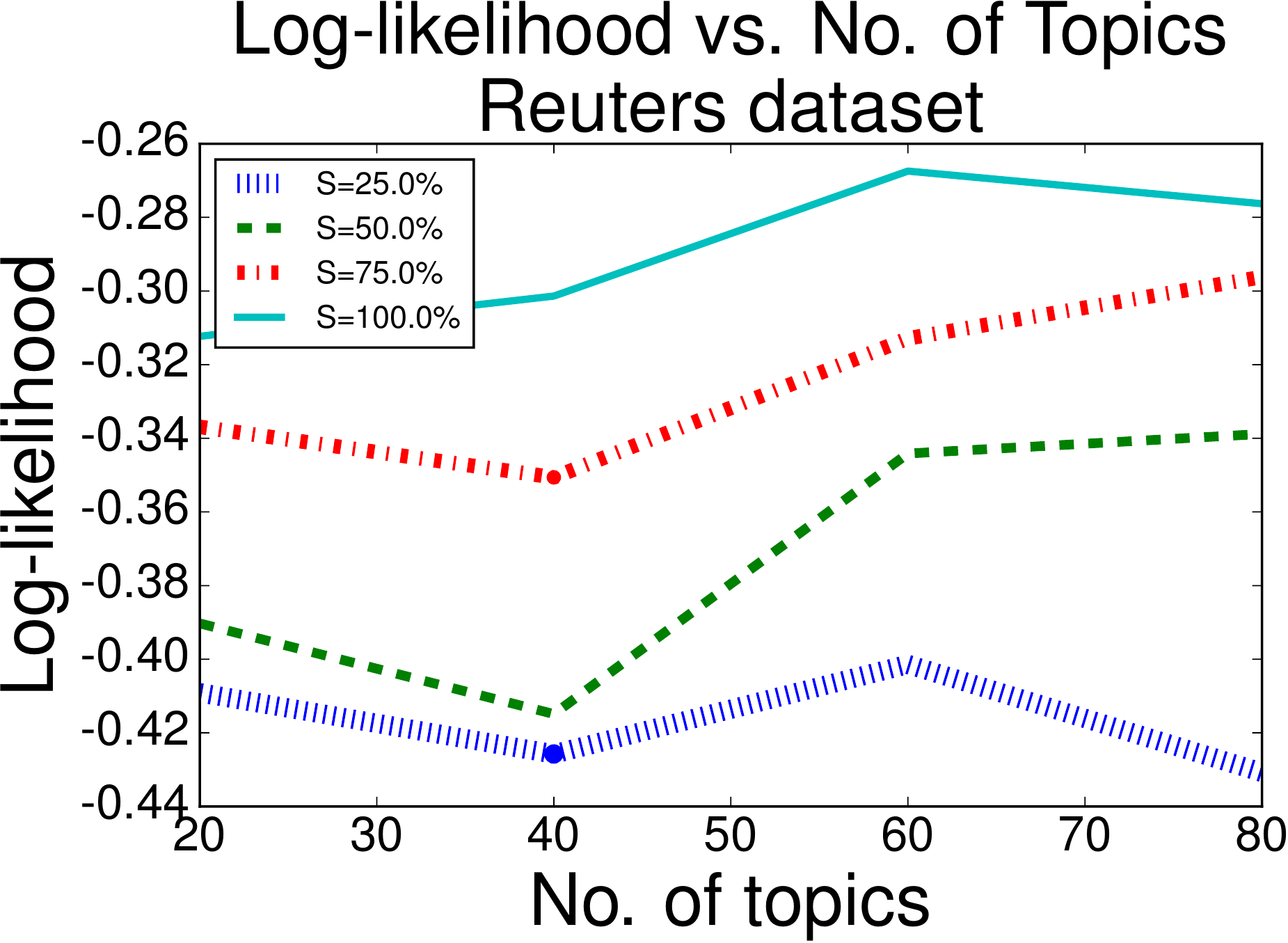}
\subcaption{}
\end{minipage}
\caption{Performance of ML-PA-LDA-C on Reuters dataset. T is the number of topics, C is the number of classes and S is the percentage of dataset used for training. An increase in the number of the training set favors an improvement in the various measures. Similarly increase in the number of topics favors an improvement in the measures provided sufficient training data is available.  Other datasets (omitted for space) follow a similar trend.}
\label{fig:ann-topics-dataset-size}
\end{figure}

Similar to the experiment carried out on the non-annotator model, we vary the number of topics as well as data-set sizes and compute all the measures used. The plots are shown  in \Cref{fig:ann-topics-dataset-size}. As in the non-annotator model, an increase in the topics as well as dataset size improves the performance of the algorithm in general. The Occam's razor observation holds here too as similar performane is achieved for a small as well as large number of topics when the size of the training dataset is small. But as more training data becomes available, having more number of topics helps.

\subsubsection{Adversarial Annotators:}
We also tested the robustness of our model against labels from adversarial or malicious annotators. An adversarial annotator is characterised by a quality parameter $\rho < 0.5$. As in the previous case, we simulated 50 annotators. The $\rho$ values of 10 of them was sampled from $U[0.0001, 0.1]$. For another 15 annotators, $\rho$ was sampled from $U[0.51, 0.65]$. For another 20 of them $\rho$ was sampled from  $U[0.66, 0.85	]$ and for the remaining 5 of them $\rho$ was sampled from $U[0.86, 0.9999]$. On the Reuters dataset, we obtained  an average accuracy of \textbf{0.955}, average class log likelihood of \textbf{-0.193}, average micro-f1 of \textbf{0.793} and an average ann-rmse of	 \textbf{0.002} over five runs, with 40 topics. This shows that even in the presence of malicious annotators, our model remains unaffected and performs well.

\section{Conclusion}
We have introduced a new approach for multi-label classification using a novel topic model, which uses information about the presence as well as absence of classes. In the scenario when the true labels are not available and instead a noisy version of the labels is provided by the annotators, we have adapted our topic model to learn the parameters including the qualities of the annotators. Our experiments on real world datasets demonstrate the superior performance of our approach.

\bibliographystyle{abbrv}
\bibliography{ml-crowd}
\end{document}